\documentclass[conference]{IEEEtran}
\IEEEoverridecommandlockouts
\usepackage{cite}
\usepackage{amsmath,amssymb,amsfonts}
\usepackage{algorithmic}
\usepackage{graphicx}
\usepackage{textcomp}
\usepackage{xcolor}
\usepackage{pifont}
\usepackage{subcaption}
\usepackage{hyperref}
\def\BibTeX{{\rm B\kern-.05em{\sc i\kern-.025em b}\kern-.08em
    T\kern-.1667em\lower.7ex\hbox{E}\kern-.125emX}}

\usepackage{fancyhdr}
\usepackage{kantlipsum}
\fancypagestyle{plain}{
\fancyhf{}
\fancyhead[C]{Conference on \LaTeX} 

}
\usepackage{eso-pic}

\begin{document}
\title{Noise Audits Improve Moral Foundation Classification}

\author{\IEEEauthorblockN{Negar Mokhberian\textsuperscript{\ding{170}}}
\and
\IEEEauthorblockN{Frederic R. Hopp\textsuperscript{\ding{169}}}
\and
\IEEEauthorblockN{Bahareh Harandizadeh\textsuperscript{\ding{170}}}
\and
\IEEEauthorblockN{Fred Morstatter\textsuperscript{\ding{170}}}
\and
\IEEEauthorblockN{Kristina Lerman\textsuperscript{\ding{170}}}
\and
\IEEEauthorblockA{\textsuperscript{\ding{170}}\textit{Information Sciences Institute,} \\
\textit{University of Southern California}\\
\{nmokhber, harandiz, fredmors, lerman\}@isi.edu}
\and 
\IEEEauthorblockA{\textsuperscript{\ding{169}}\textit{Amsterdam School of Communication Research,} \\
\textit{University of Amsterdam}\\
f.r.hopp@uva.nl}
}

\maketitle

\begin{abstract}
Morality plays an important role in culture, identity, and emotion. Recent advances in natural language processing have shown that it is possible to classify moral values expressed in text at scale. Morality classification relies on human annotators to label the moral expressions in text, which provides training data to achieve state-of-the-art performance. However, these annotations are inherently subjective and some of the instances are hard to classify, resulting in noisy annotations due to error or lack of agreement. The presence of noise in training data harms the classifier's ability to accurately recognize moral foundations from text.
We propose two metrics to audit the noise of annotations. The first metric is \textit{entropy} of instance labels, which is a proxy measure of annotator disagreement about how the instance should be labeled. The second metric is the \textit{silhouette} coefficient of a label assigned by an annotator to an instance. This metric leverages the idea that instances with the same label should have similar latent representations, and deviations from collective judgments are indicative of errors. Our experiments on three widely used moral foundations datasets show that removing noisy annotations based on the proposed metrics improves classification performance.\footnote{Our code can be found at: \url{https://github.com/negar-mokhberian/noise-audits}} 
\end{abstract}


\section{Introduction}



{{Moral foundations theory (MFT)}}~\cite{graham2009liberals, graham2013moral} suggests that the moral values expressed in opinions, thoughts, and cultures 
can be explained by five universal, but contextually variable \textit{moral foundations}. These foundations are typically described along bipolar dimensions: care vs. harm, fairness vs. cheating, authority vs. subversion, loyalty vs. betrayal, and purity vs. degradation. MFT was first introduced in social psychology and has found many applications in political science and the social sciences. For example, moral foundations motivate behaviors such as charitable donations \cite{hoover2018moral}, violent protests \cite{mooijman2018moralization} and social homophily \cite{dehghani-2016-purity}.


The broad adoption of MFT was driven, in part, by advances in natural language processing (NLP), which enabled researchers to quantify moral values expressed in text, including news~\cite{mokhberian2020moral, hopp2021extended, weber2018extracting}, political speech~\cite{wang2021moral}, and social media discussions~\cite{hoover2020moral}, at scale. Early works relied on lexicons that defined words associated with moral virtues and vices to classify moral foundations from text~\cite{graham2009liberals}. However, by neglecting semantic context in sentences, lexicon-based approaches fail to capture the nuances of moral expression~\cite{hopp2021reflections}. To address this challenge, more recent approaches use large language models to capture the moral context of text~\cite{kennedy2021moral, xie2020contextualized}. These approaches leverage a text corpus manually annotated for moral values to train language models to recognize examples of moral language. Several such ground truth data sets exist~\cite{hoover2020moral, hopp2021extended,
trager2022moral,
johnson2017modeling, weber2018extracting}.

\begin{figure}[ht!]
\begin{subfigure}{.45\textwidth}
  \centering
  \includegraphics[width=\linewidth]{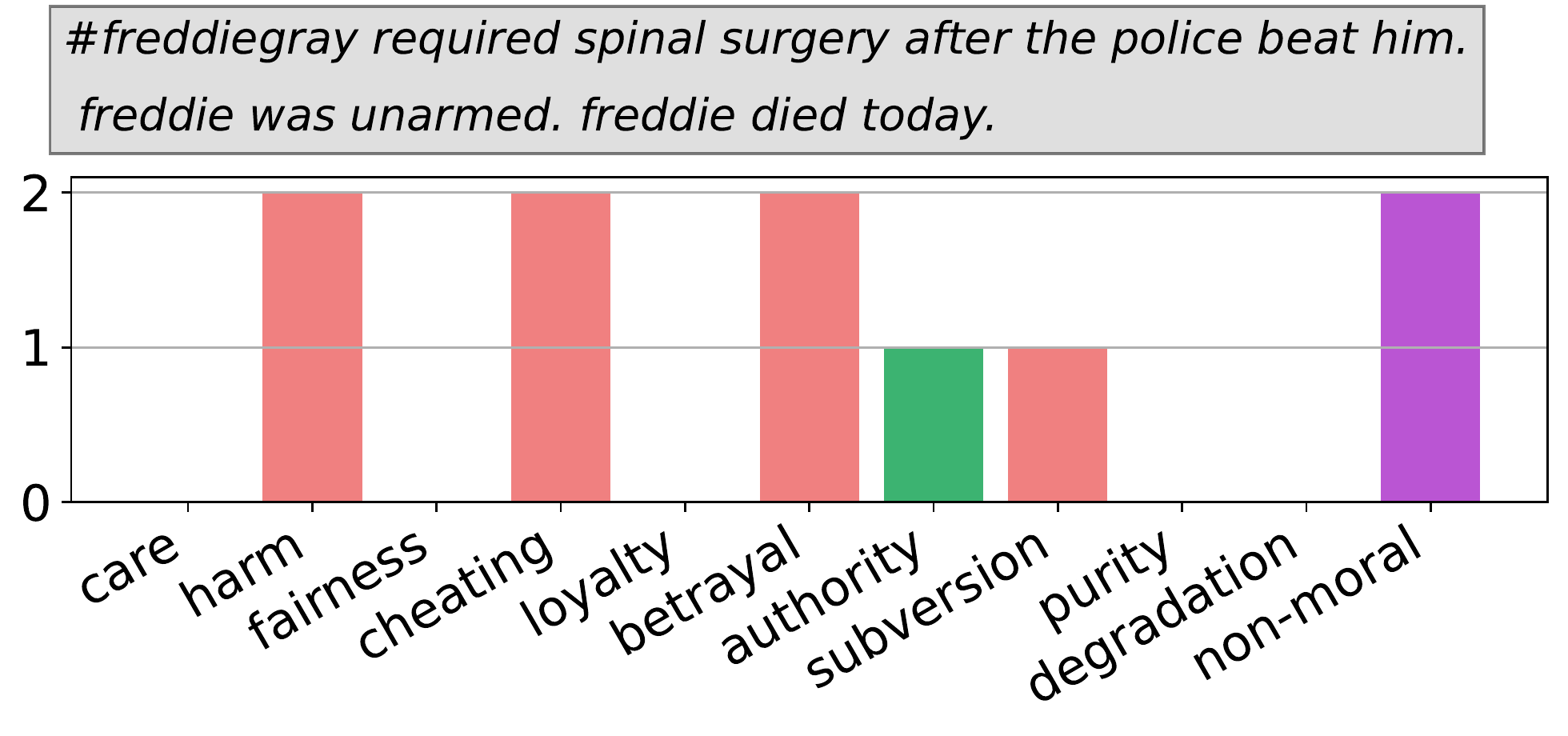}  
  \caption{This tweet organically contains different  dimensions of morality. It contains elements of injury and police action at the same time.}
  \label{fig:sub-first}
\end{subfigure}
\begin{subfigure}{.45\textwidth}
  \centering
  \includegraphics[width=\linewidth]{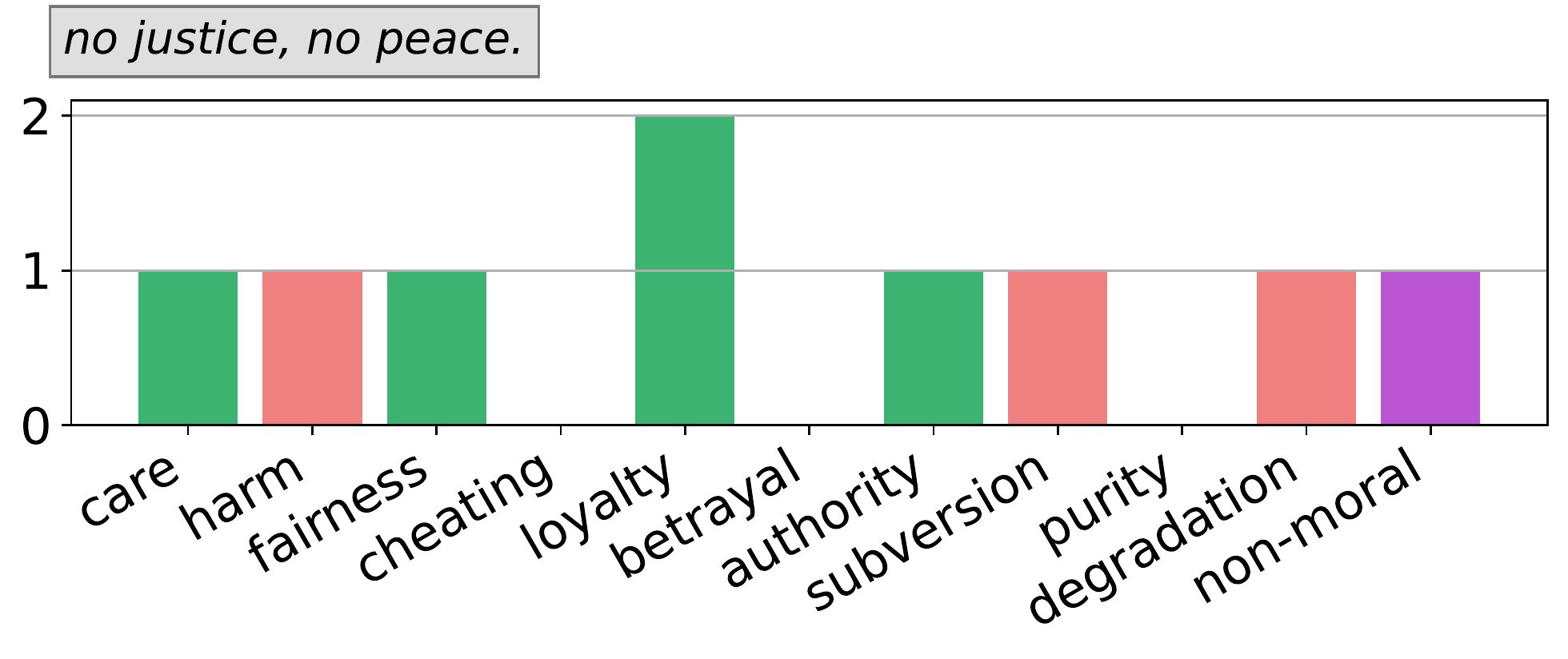}  
  \caption{This instance has been annotated with ``degradation", and ``non-moral" but these labels are not related to it.}
  \label{fig:sub-second}
\end{subfigure}
\begin{subfigure}{.45\textwidth}
  \centering
  \includegraphics[width=\linewidth]{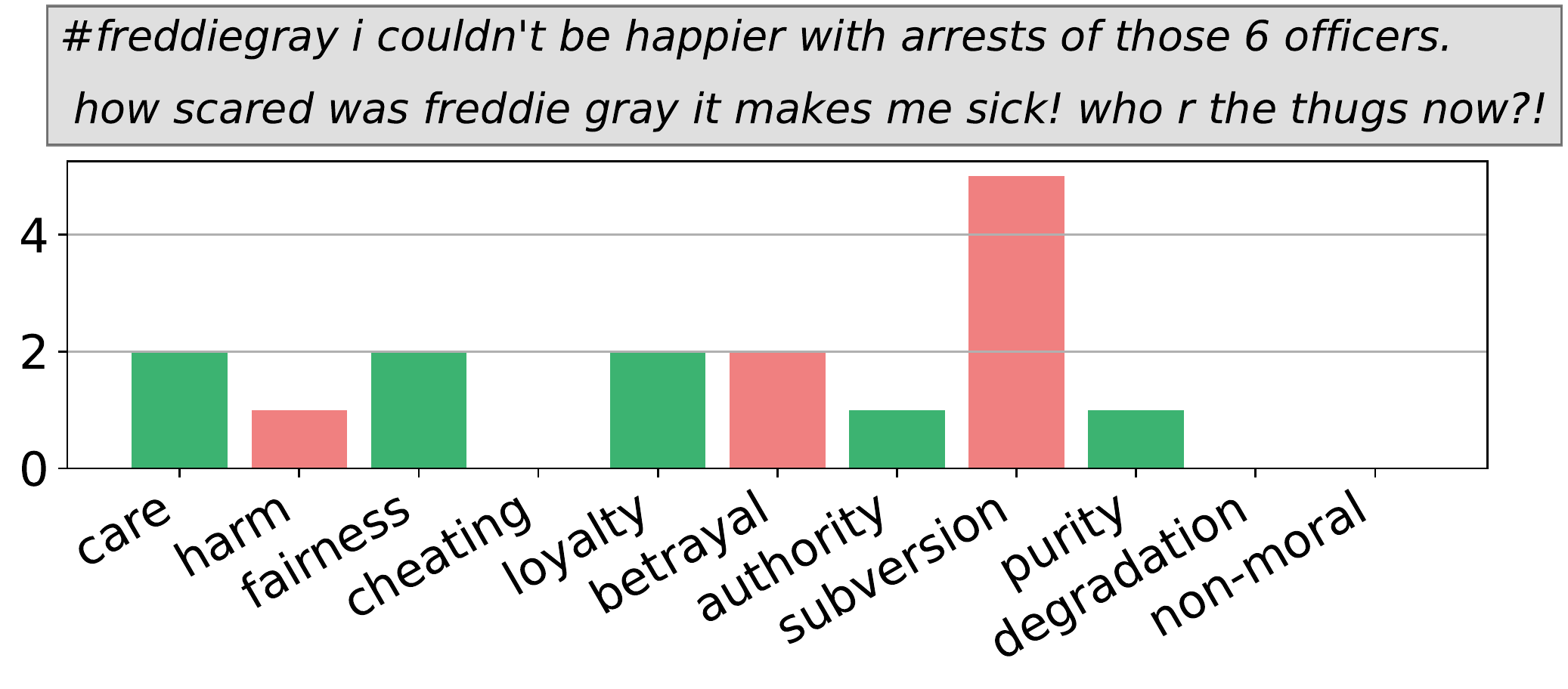}  
  \caption{This instance's moral labels are subjective, depending on whether the annotators support the police actions or are against it.}
  \label{fig:sub-third}
\end{subfigure}
\caption{Examples of high-entropy instances in MFTC dataset. The text of the tweet and the number of annotators selecting each moral foundation as the label are shown.}
\label{fig:examples}
\end{figure}

A major challenge when constructing ground truth data for training moral foundation classifiers is the subjectivity of individual moral judgments, which are prone to bias and noise~\cite{kahneman2021noise}. Figure~\ref{fig:examples} illustrates this challenge using real examples from the Moral Foundation Twitter Corpus (MFTC)~\cite{hoover2020moral}.
The first tweet, ``\#freddiegray required spinal surgery after the police beat him. freddie was unarmed. freddie died today,'' expresses a range of moral concerns (Fig. \ref{fig:sub-first}). It was labeled as \textit{harm}, \textit{cheating}, \textit{betrayal} and \textit{non-moral} by two annotators each, and \textit{authority} and \textit{subversion} by one annotator each. Text can also be mislabeled due to errors or ambiguity. This is illustrated by the second example ``no justice, no peace.'' (Fig. \ref{fig:sub-second}). This instance was annotated with \textit{degradation}, and \textit{non-moral}, but these labels are not related to the tweet. The third example illustrates the subjectivity of moral judgments (Fig. \ref{fig:sub-third}): ``\#freddiegray I couldn't be happier with arrests of those 6 officers. how scared was freddie gray it makes me sick! who r the thugs now?!'' The moral labels assigned to this instance are subjective and depend on whether the annotators support police actions or not.
As a result of these factors, individual moral labels will be noisy. To partly control for the variability of judgments, researchers use the label chosen by the majority of annotators as the correct ground truth label for each instance~\cite{hoover2020moral,prabhakaran-etal-2021-releasing}. However, the question of how annotation noise affects the performance of models learned from data  has been underexplored~\cite{davani2022dealing,swayamdipta-etal-2020-dataset}. 

In this paper we present and compare two approaches for auditing and removing noise from annotations used to train moral foundations classifiers.
The first approach identifies difficult instances in the ground truth data. We propose \textit{entropy} as a measure of annotator disagreement in \S\ref{sec:entropy} and use it to identify instances with little agreement that will degrade classifier training. For each instance, we calculate \textit{entropy} based on how many annotators have selected each of the labels (examples of such distributions are shown in Fig.~\ref{fig:examples} for three instances). By removing instances on which the annotators disagree, we hope to create better data for training classifiers.

Our second method identifies annotations that deviate from collective judgments of all annotators. We propose the \textit{silhouette} coefficient in latent space as a measure of label quality, which leverages the idea that similarly-labeled instances should have similar latent representations. By removing annotations that deviate substantially from collective judgments, we hope to improve the quality of ground truth data. 

We evaluate both approaches on three large datasets with moral annotations~\cite{hoover2020moral, hopp2021extended, trager2022moral} (more information of the datasets in \S\ref{sec:data}). 
We show that training models on ground truth data from which noisiest annotations have been removed improves morality classification.
%
%
This does not stem simply from having less data: compared to a model trained on data from which the same number of instances were removed at random, removing the noisiest instances from data significantly improves classification performance.

Our work primarily focuses on noise audit of subjective annotations to identify mislabeled or difficult instances in moral foundations classification. This could be applied in any subjective labeling setting. Identifying moral foundations in text is a representative example of a subjective task that we have selected for this paper.
An important enabler for our noise audit is access to the individual judgments of annotators. Hence, we encourage the crowd-sourced annotation builders to include fine-grained details of individual annotator judgments on instances instead of the common practice \cite{sabou2014corpus} of reporting an aggregated judgment (e.g., majority label). Including the individual judgments gives the opportunity to refine the dataset, enhancing its utility on learning tasks.

\section{Related Work}
{Researchers have tried to incorporate  individual annotator's perspectives of the subjective tasks~\cite{davani2022dealing}. Using multi-annotator models, they have shown that training a separate model for each annotator and then aggregating to a majority vote performs better than aggregating labels in the data prior to training. However, their methodology is not practical in the cases where the data have been crowd-sourced and there are many annotators because 1) there are not usually enough data points per annotator to train separate modules, and 2) it is not cost-efficient to train many separate modules.}

Other earlier works have studied the difficulty of data points through information theory \cite{ethayarajh2021information} and disagreements in annotations \cite{pavlick-kwiatkowski-2019-inherent}. A leading way to understand the difficulty of data instances is to leverage training dynamics~\cite{pleiss2020identifying, pulastya2021assessing}. In other words, by observing how a classifier performs on an specific instance through epochs. Perhaps the best exemplar of this approach is seen in~\cite{swayamdipta-etal-2020-dataset}, which assesses each point based on the model's \emph{confidence} (average probability assigned to the correct label during training epochs) and \emph{variance} (variance in probability assigned to the correct label during training epochs). While training dynamics offer a way to measure training difficulty on an instance, their main drawback is dependency on the design of the model and on training. However, we try to address this issue by proposing metrics that are only depended on the annotated dataset and training is not a requirement for their calculation. We apply our proposed metrics to refine moral foundations datasets before training. To the best of our knowledge no prior work has analysed how difficult instances of moral foundations affect classification.

\begin{figure*}[ht!]
    \begin{subfigure}{0.48\textwidth}
        \centering
        \includegraphics[width=\textwidth]{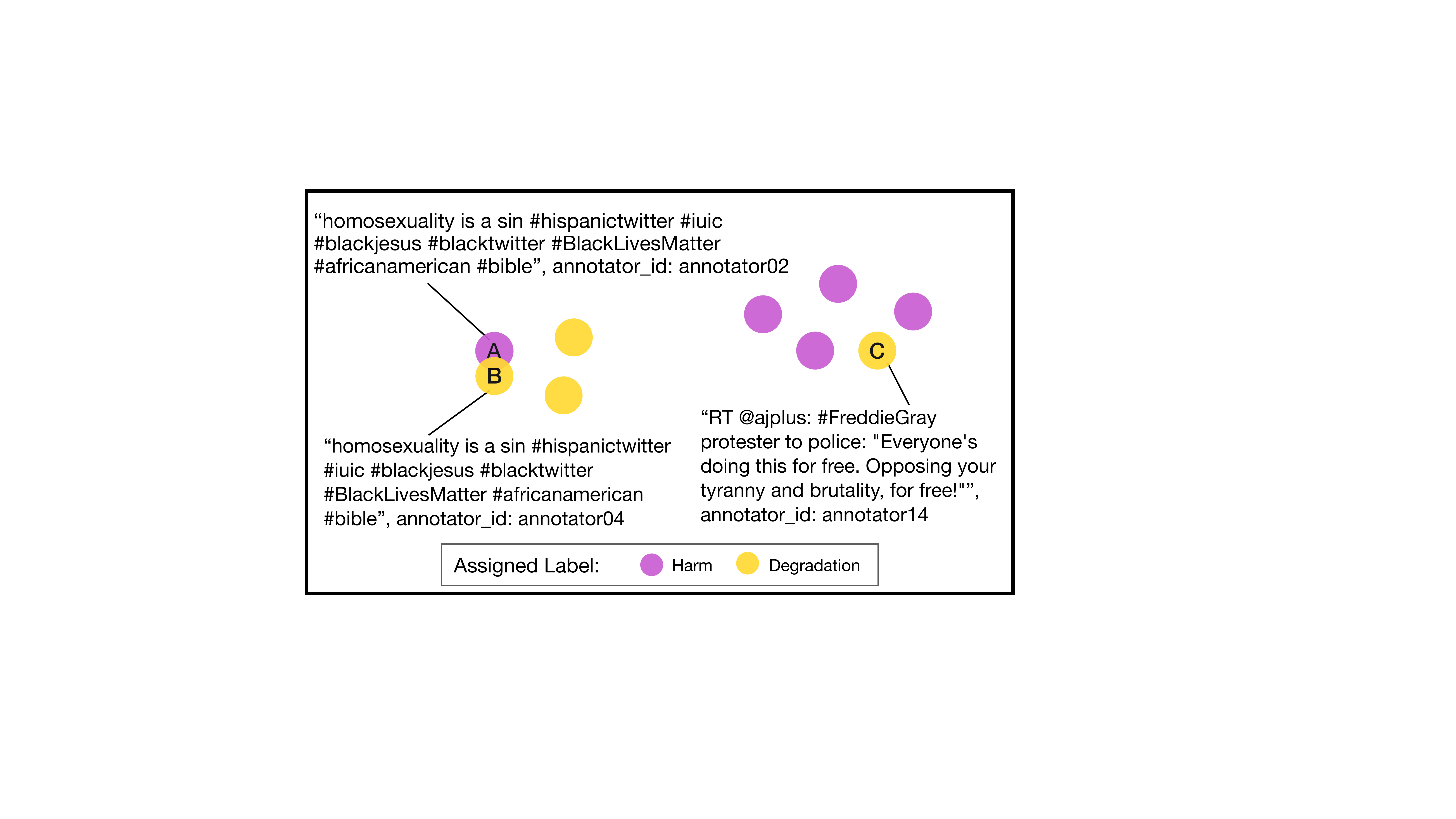}
        \caption{MFTC dataset}
        \label{fig:sil_mftc}
    \end{subfigure}
     \begin{subfigure}{0.48\textwidth}
        \centering
        \includegraphics[width=\textwidth]{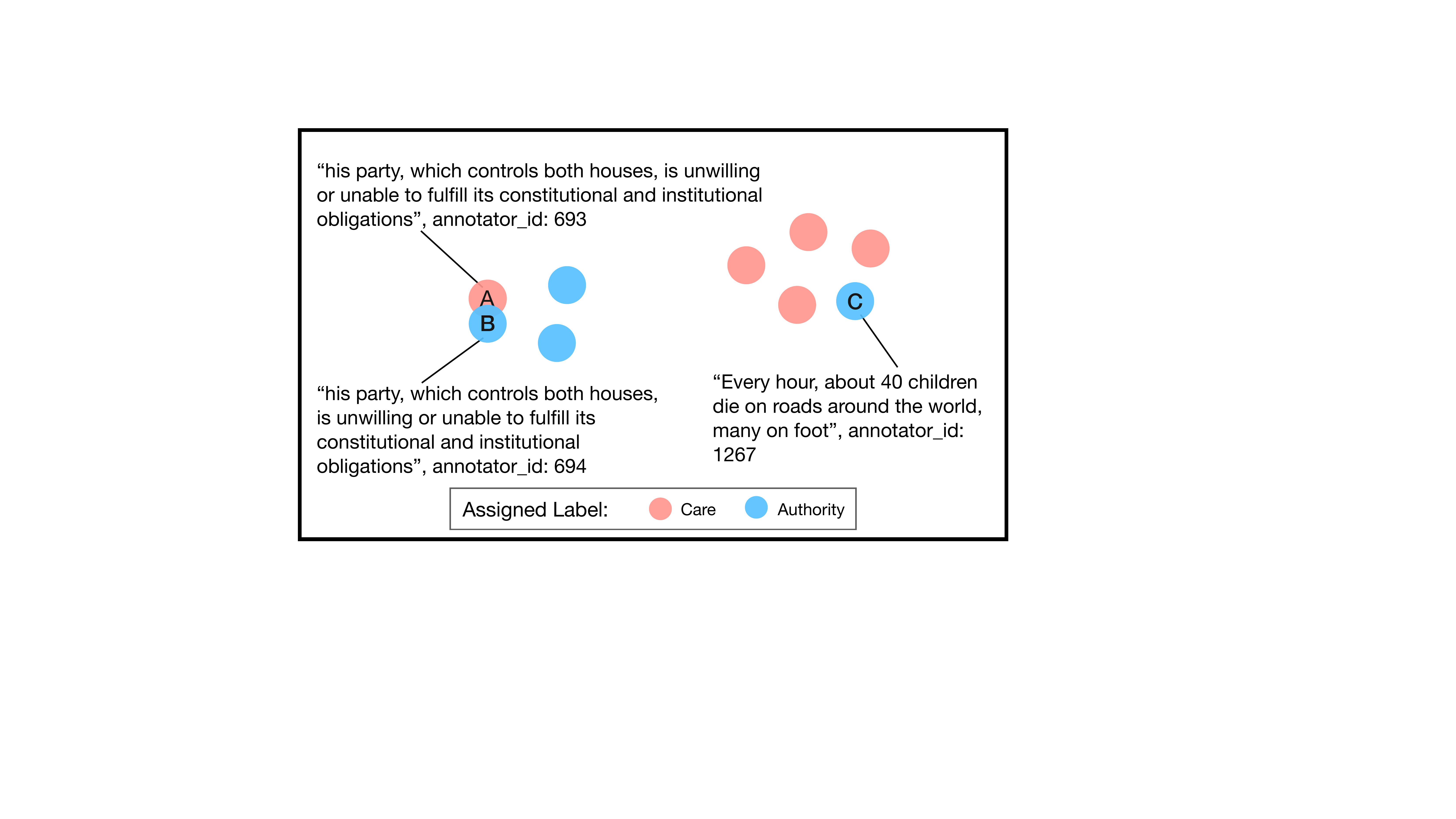}
        \caption{MFNC dataset}
        \label{fig:sil_emfd}
    \end{subfigure}
    \caption{Examples of moral judgments with low silhouette coefficients. Judgments A, B are on the same text but from different annotators. In the language model latent space this text is very close to other judgments of \textit{Degradation} in the left figure and to \textit{Authority} in the right figure. In judgment A the annotator has assigned it a label that does not match the labels of similar texts. Because it is far from other texts with the same label, but close to texts that have been assigned different labels, judgment A will get a low silhouette coefficient.  Judgment B is close to the other judgments of the same label so it will get high silhouette coefficient. Our methodology suggests removing judgment A from the training data but to keep judgment B on the same text. In judgment C the annotator has also selected a label that is different from the label of the other similar texts, so we suggest to remove judgment C.}
    \label{fig:sil_examples}
    \vspace{-5pt}
\end{figure*}

\section{Methodology}
\label{sec:method}
In this section we propose two metrics \textit{\textbf{entropy}} and \textit{\textbf{silhouette}} for identifying noise in instance-level and judgment-level respectively. We use \textit{entropy} (see \S \ref{sec:entropy}) to find out annotator disagreements on a given instance. An instance is a piece of text that has been shown to annotators to collect their judgments. The disagreement on an instance is high when disparate judgments have been collected from annotators; on the other hand, the disagreement is low when all annotators agree on the same label. We refine the datasets by removing the high-entropy instances (instances with high disagreement on their assigned labels) as a pre-processing step before training a classifier.

In addition, we propose using the \textit{silhouette} coefficient (see \S \ref{sec:sil}) as a fine-grained metric to quantify how a single judgment differs from other instances of the same-label judgments. 
Note that unlike entropy, this metric takes into account the text of the instance.
Removing judgments we deem noisy with this metric increases the inter-annotator agreement on a given instance. 

Unlike prior work in machine learning \cite{pleiss2020identifying, toneva2018an, rodriguez-etal-2021-evaluation, swayamdipta-etal-2020-dataset}, 
our measures are calculated before the model training starts and do not depend on the training dynamics. {We provide the details of one the training dynamics metrics in \S\ref{sec:cartography} and in the experiments (\S\ref{sec:results}) will monitor their improvement as we filter data based on our metrics.}

\subsection{Entropy at the Instance Level}
\label{sec:entropy}
We use \textit{entropy of annotations} to quantify diversity of labels gathered for a text. For a text $t_i$ and its multi-label annotations $<l_1:c_{i1}, ...,l_{N}:c_{iN}>$ in which $l_j$ is a member of all the labels $L=\{l_1,l_2, ..., l_N\}$ and $c_{ij}$ is the count of annotators who have assigned $l_j$ to $t_i$, we calculate the \textit{entropy} as:
$$entropy(t_i) =- \sum_{j=1}^{N}{P(l_j, t_i)log P(l_j, t_i)}$$
in which 
$$P(l_j, t_i)= \frac{c_{ij}}{\sum_{j=1}^{N} c_{ij}}.$$
If all annotators agree on the same label, the \textit{entropy} is zero. At the other extreme, if every annotator gave the instance a different label, entropy has its maximum value. 


\subsection{Silhouette Coefficient at the Judgment Level}
\label{sec:sil}
For a given judgment $x$ = ($t_i$, $l_j$, $a_k$) in which instance $t_i$ has been assigned a label $l_j$ by annotator $a_k$, the \textit{silhouette} coefficient \cite{rousseeuw1987silhouettes} is defined as a combination of its distance from the same-label (intra-cluster) judgments and from other-label (inter-cluster) judgments. 
Let $l_j$ be a member of the set of all labels $L=\{l_1,l_2, ..., l_N\}$. We consider all the instances assigned to the same label to be a cluster. The intra-cluster measure  $a(x)$ is defined as the average dissimilarity of $text_i$ to all other texts labeled with $l_j$. The inter-cluster metric is defined as: 
 $$b(x) = min_{y\in L, y\neq l_j} \bar{d}(t_i, T_y),$$
 where  $\bar{d}(t_i, T_y) $ is the average dissimilarity of $t_i$ to all texts in the dataset labeled with $y$. Finally, $a(x)$ and $b(x)$ are aggregated as: 
$$silhouette(x)=\frac{b(x)-a(x)}{max\{a(x),b(x)\}}$$
To calculate the dissimilarities ($d$), we represent the text instances in a latent space using a language model and calculate the distance of vectors corresponding to texts. 

The \textit{silhouette} coefficient thus captures the consistency of an annotator's label of a specific text with other same-label texts. If the instance is too different from the content of other examples with the same label, it will have a low \textit{silhouette} coefficient, and we consider it to be noise. Training the model on this judgment may confuse it and reduce classification performance. Hence, we suggest removing this judgment from the ground truth corpus before training. However, other high-silhouette judgements for $t_i$ may be preserved and $t_i$ can be part of the filtered dataset using other judgements of it. 

\subsection{Model's Training Dynamics}
\label{sec:cartography}
Recent work by \cite{swayamdipta-etal-2020-dataset} uses signals from epochs during training (training dynamics) as a proxy for exploratory data quality estimation. 
They define conﬁdence of the model on instance $i$ as the average of model's probability of its true label ($y_i^*$) across epochs:

$${\hat\mu}_i = \frac{1}{E}\sum_{e=1}^{E} p_{\theta^{(e)}}(y_i^*|t_i)$$
where $p_{\theta^{(e)}}$ indicates to the model’s probability with parameters $\theta^{(e)}$ at the end of the epoch $e$. Their experiments show that the examples with low confidence are likely to be mislabeled. 
\section{Experiments}
\subsection{Data}
\label{sec:data}

\begin{figure*}[ht!]
\begin{subfigure}{\textwidth}
    \begin{subfigure}{0.5\textwidth}
        \centering
        \includegraphics[width=\textwidth]{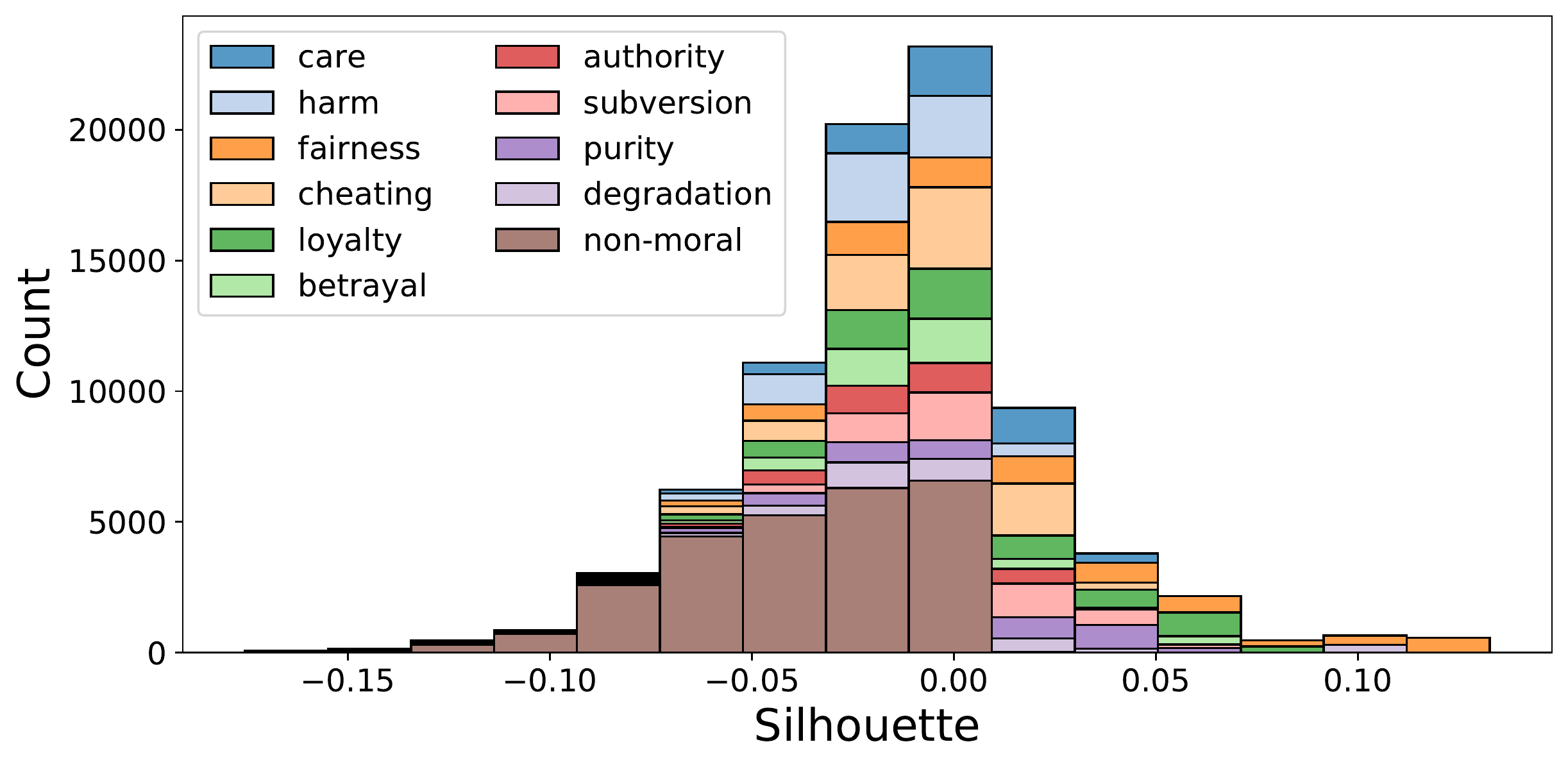}
        \label{fig:category_sil_mftc}
    \end{subfigure}
    \begin{subfigure}{0.5\textwidth}
        \centering
        \includegraphics[width=\textwidth]{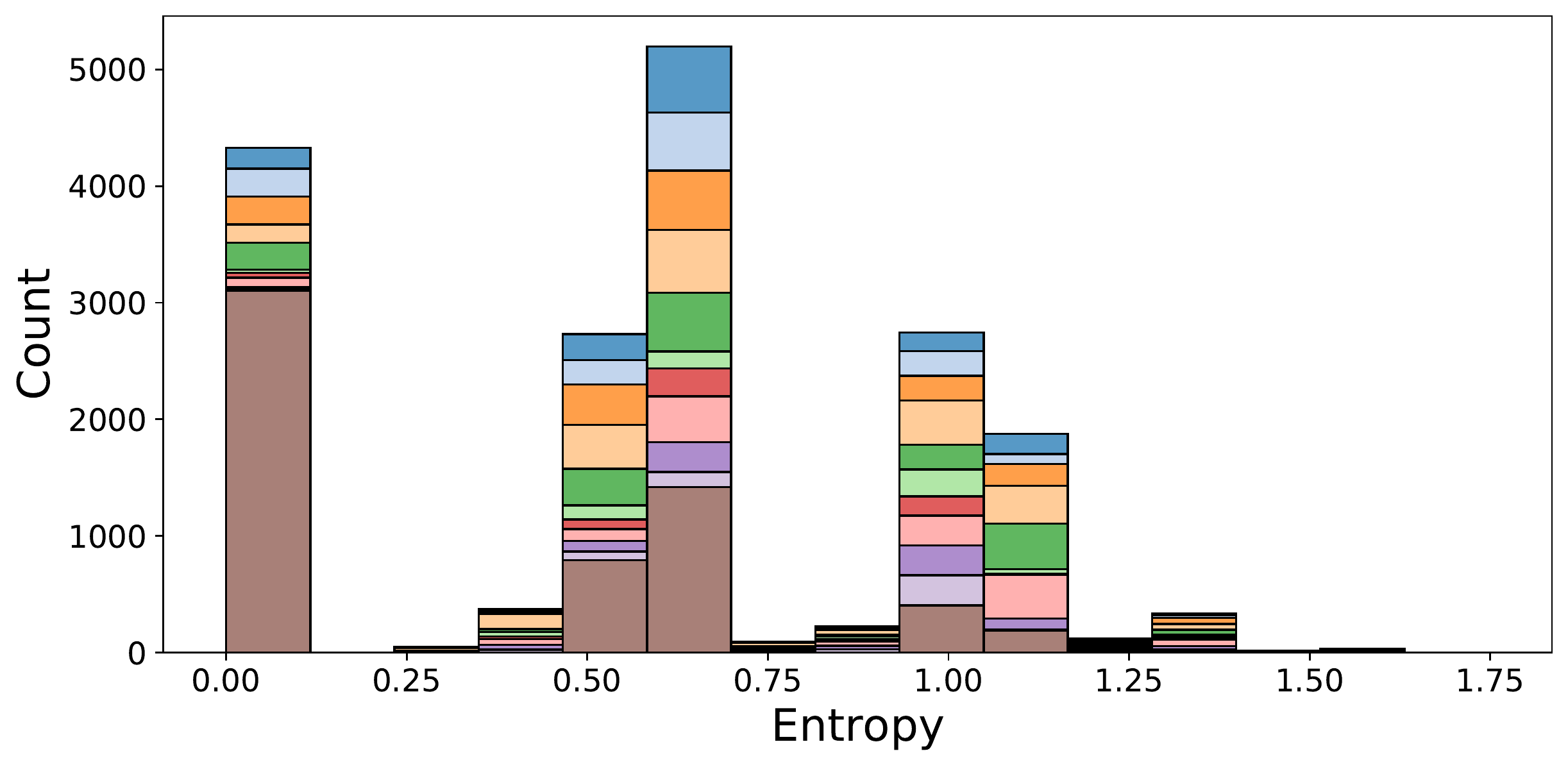}
        \label{fig:category_entropy_mftc}
    \end{subfigure}
\caption{Moral Foundation Twitter Corpus (MFTC)}
\end{subfigure}

\begin{subfigure}{\textwidth}
    \begin{subfigure}{0.5\textwidth}
        \centering
        \includegraphics[width=\textwidth]{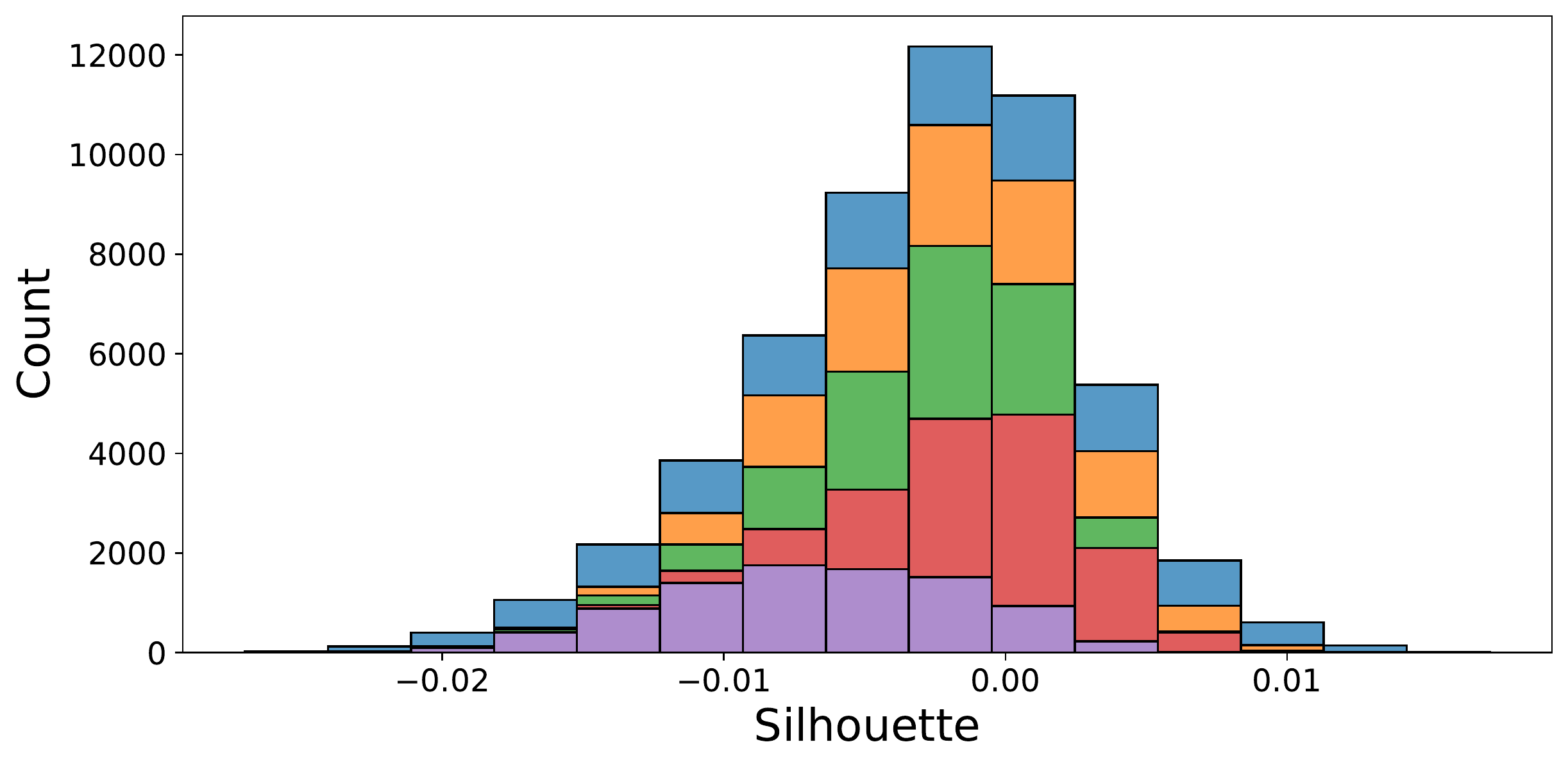}
        \label{fig:category_sil_emfd}
    \end{subfigure}
     \begin{subfigure}{0.5\textwidth}
        \centering
        \includegraphics[width=\textwidth]{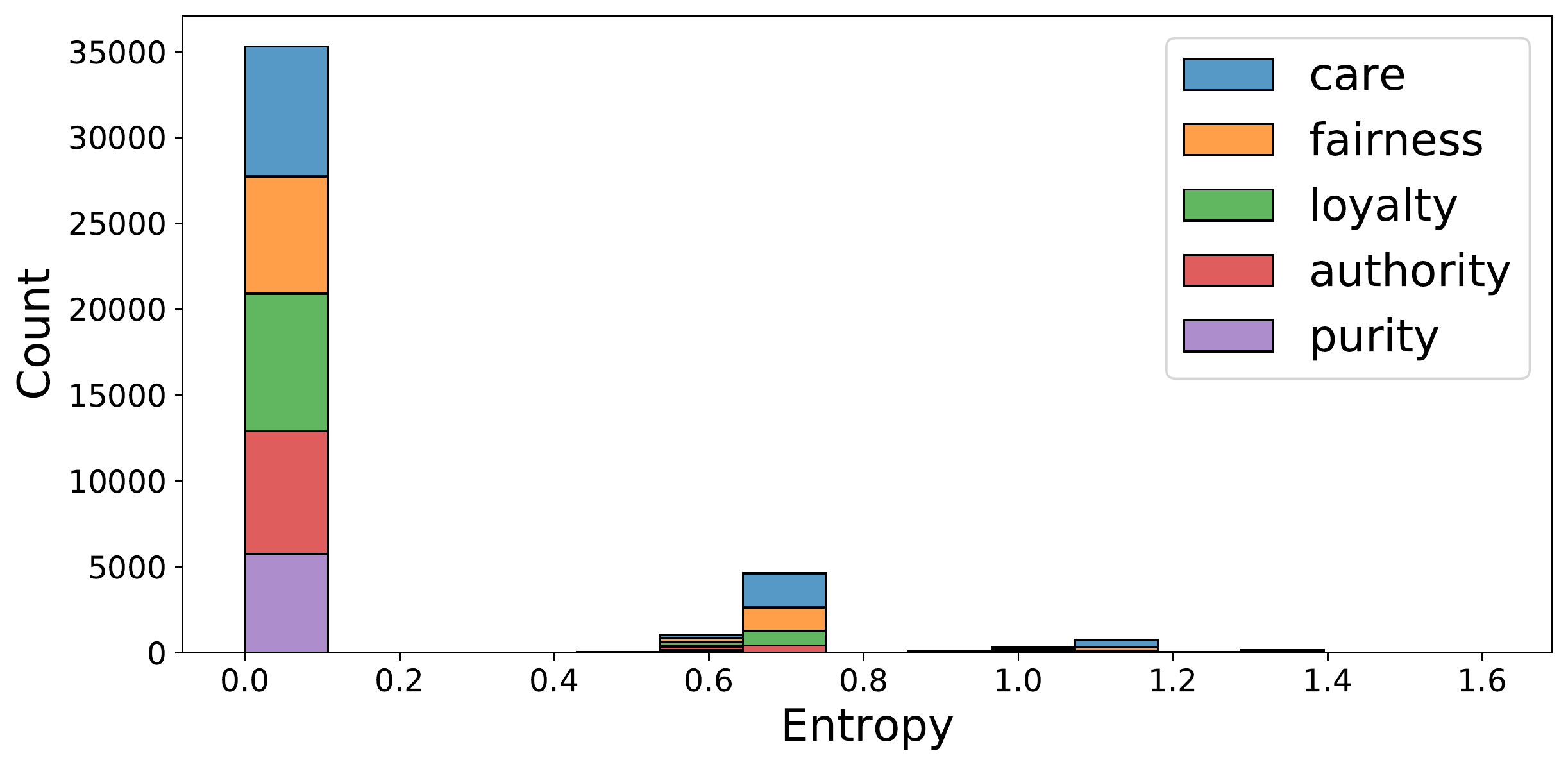}
        \label{fig:category_entropy_emfd}
    \end{subfigure}
\caption{Moral Foundation News Corpus (MFNC)}
\end{subfigure}
\begin{subfigure}{\textwidth}
    \begin{subfigure}{0.5\textwidth}
        \centering
        \includegraphics[width=\textwidth]{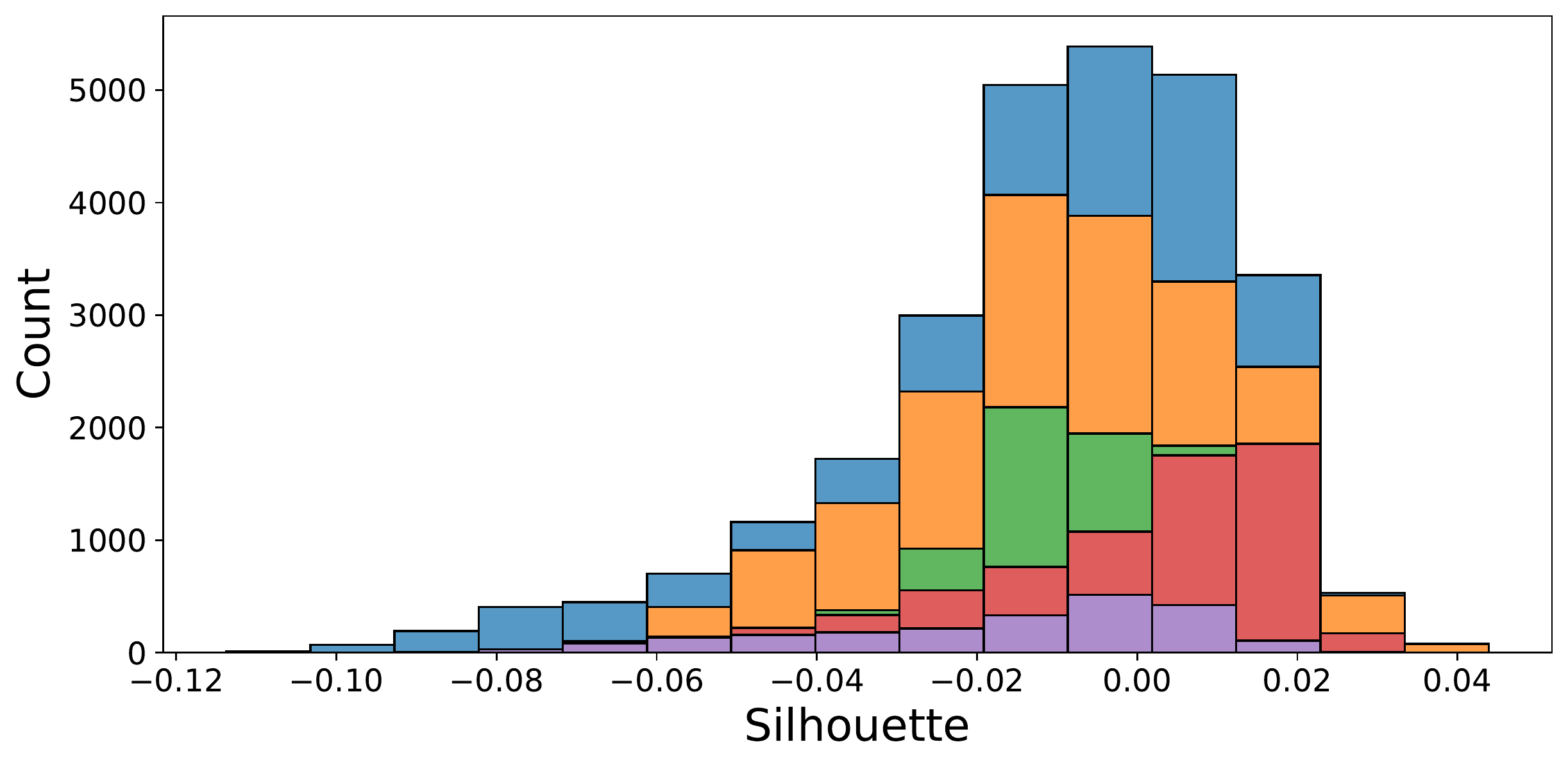}
        \label{fig:category_sil_mfrc}
    \end{subfigure}
     \begin{subfigure}{0.5\textwidth}
        \centering
        \includegraphics[width=\textwidth]{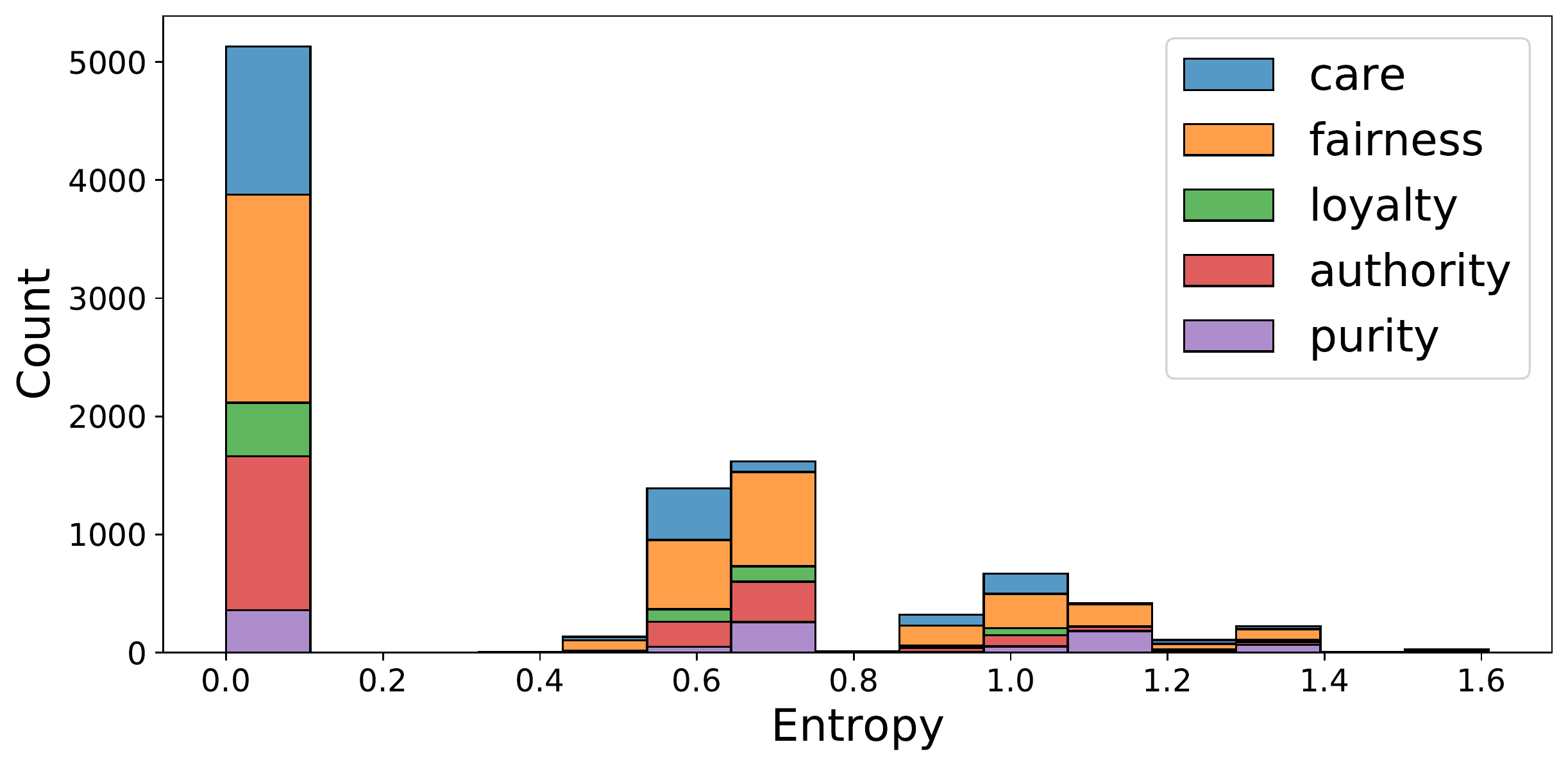}
        \label{fig:category_entropy_mfrc}
    \end{subfigure}
\caption{Moral Foundation Reddit Corpus (MFRC)}
\end{subfigure}
    \caption{Measures of disagreement. The left column shows the distributions of the silhouette metric for the three datasets. The silhouette coefficient is calculated on annotations. 
    The  right column show the distributions of entropy metric for instances, colored by the majority-vote of moral foundation assigned to the instance.}
    \vspace{-5pt}
    \label{fig:datasets_metrics}
\end{figure*}

    

In this paper we focus on three large moral foundations annotated textual datasets. In these datasets the annotations of many texts are very diverse. This diversity can appear because of the difficulty of the labeling task, subjectivity of moral judgments, the bias based on personal beliefs, ambiguity of texts, and errors made by annotators.

\subsubsection{The Moral Foundations Twitter Corpus (MFTC)~\cite{hoover2020moral}}
\textbf{\emph{The Moral Foundations Twitter Corpus (MFTC)}} is a textual multi-label dataset containing tens of thousands of tweets related to various social movements, with each tweet annotated with categories of moral foundations. For annotating MFTC, several human annotators were trained to manually annotate 35k tweets. The tweets are drawn from 7 socially relevant topics: All Lives Matter (ALM), Black Lives Matter (BLM), the Baltimore protests, the 2016 Presidential election, hate speech and offensive language, Hurricane Sandy, and \#MeToo. For each tweet several annotators selected as many moral foundations as they saw relevant, which resulted in a multi-label dataset. Note that for each tweet, the number of annotators that selected each class is known. There are in total eleven labels (ten moral foundations and one ``non-moral'' class indicating the text does not have moral relevance).

\subsubsection{The Moral Foundations News Corpus (MFNC)~\cite{weber2018extracting}}
In the MFNC ~\cite{weber2018extracting}, a crowd of 854 annotators was drawn from the general United States population using the crowd-sourcing platform Prolific Academic (PA\footnote{\url{https://www.prolific.ac/}}). Sampling was designed to match annotator characteristics to the US general population in terms of political affiliation and gender, thereby lowering the likelihood of obtaining annotations that reflect the moral intuitions of only a small, homogeneous group (see supplemental materials in \cite{hopp2021extended} for detailed information on annotators) . Fifteen randomly selected news documents (from among 2,995 articles total) were assigned to each annotator. The selected corpus consisted of online newspaper articles discussing a wide range of sociopolitical topics from 11 prominent, U.S. news outlets. 557 annotators completed all assigned tasks. Each annotator underwent an online training explaining the purpose of the study, the basic tenets of MFT, and the annotation procedure. Annotators were instructed that they would be annotating news articles, and that for each article they would be (randomly) assigned \textit{one} of the five moral foundations. Next, using a digital highlighting tool, annotators were instructed to highlight all portions of a news article that they understood to reflect their assigned moral foundation. In total, 63,958 annotations (i.e., textual highlights) were produced by the 557 annotators. Note that the coding task of the MFNC differed from the MFTC task in important ways: First, annotators were assigned to focus on the presence of \textit{one} (randomly assigned) moral foundation per article, rather than assigning portions of the article to \textit{any} moral foundation. Second, annotators labeled the holistic presence of a moral foundation rather than differentiating whether a foundation was upheld (e.g. care) or violated (e.g. harm). Third, annotators were free to highlight \textit{portions} of a news article, in contrast to labeling the entire coding unit with a moral foundation. 

\subsubsection{The Moral Foundations Reddit Corpus (MFRC)~\cite{trager2022moral}}

The MFRC consists of 16,123 Reddit comments drawn from 12 different subreddits. Every instance has been labeled by at least three annotators from a set of ﬁve trained annotators. 

\subsection{Metrics of Disagreement}
Figure~\ref{fig:datasets_metrics} shows the distributions of \textit{entropy} and \textit{silhouette} metrics for the three datasets. The distribution varies for each category and each dataset. For example, in MFTC, ``non-moral'' has the higher concentration of posts with zero entropy, meaning that in many instances all the annotators agreed on the non-moral category. On the other hand the classes purity and degradation have few samples with zero entropy of annotations. Also, in MFNC, all the categories have high concentration of posts with zero entropy which is because there are many instances with single highlights in MFNC that have no overlap with other highlights. Figure~\ref{fig:datasets_metrics} also shows the distribution of silhouette coefficient of annotator judgements for each of the moral foundation categories. This metric has a consistently broad distribution for all datasets, unlike entropy.

\subsection{Experimental Settings}
\subsubsection{Classifier Model} Getting a text as input, our model's task is to classify it to a moral foundation label. The classification task aims to predict majority-vote for each textual instance. Which is the label with highest number of annotators selecting it for the instance (we select randomly if there is a tie). In MFTC there are eleven labels (the labels from vices and virtues of the five moral foundations plus one label denoting non-moral category). In MFNC there are only five labels related to the main foundations regardless of the polarity. For MFRC we convert the labels ``proportionality" and ``equality" to fairness and only keep the labels from the main foundations to keep consistency with MFNC.  

We use pretrained language model RoBERTa \cite{liu2019roberta} and finetune it on our datasets. This is done by adding a multi-class classiﬁcation layer on top of the language model and with a cross-entropy loss updating all the parameters end-to-end for five epochs. We minimize cross-entropy with the Adam optimizer \cite{kingma2014adam} with learning rate $2\times 10^{-5}$. Our experiments use a batch size of 50. 
We run each experiment with five random seeds and split data into $70\%$ train and $30\%$ test sets. Each experiment is performed on a single GTX 1080Ti GPU. In our implementation we use the Huggingface Transformers library \cite{wolf2019huggingface}.

\subsubsection{Silhouette Coefficient Calculation}
To calculate the silhouette coefficient of the judgments we need to use a language model to represent the text of instances in a latent space. We use sentenceBERT framework \cite{reimers2019sentence}, which has been shown to be helpful for capturing semantic textual similarity. We use the pretrained model "all-mpnet-base-v2"\footnote{\url{https://huggingface.co/sentence-transformers/all-mpnet-base-v2}}. We use the default parameters of sklearn implementation of the silhouette coefficient\footnote{\url{https://scikit-learn.org/stable/modules/generated/sklearn.metrics.silhouette_samples.html}} which uses euclidean distance as a metric to capture semantic distance of texts. 

\subsection{Results}
\label{sec:results}

\begin{figure*}[h!]
    \begin{subfigure}{\textwidth}
        \centering
        \includegraphics[width=\textwidth]{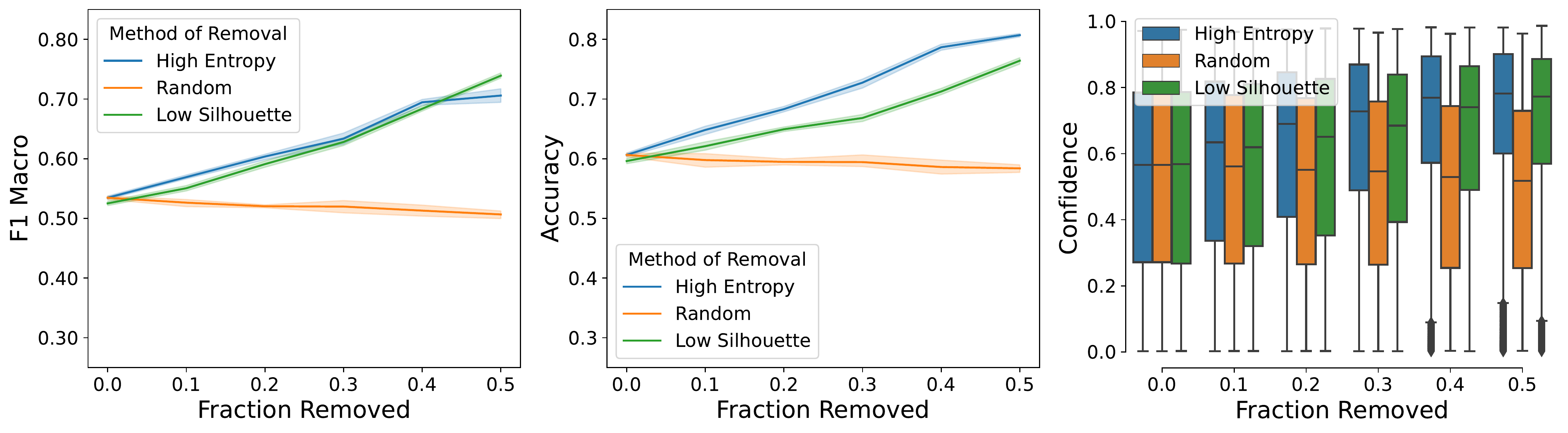}
        \caption{MFTC dataset}
        \label{fig:results_mftc}
    \end{subfigure}
    \begin{subfigure}{\textwidth}
        \centering
        \includegraphics[width=\textwidth]{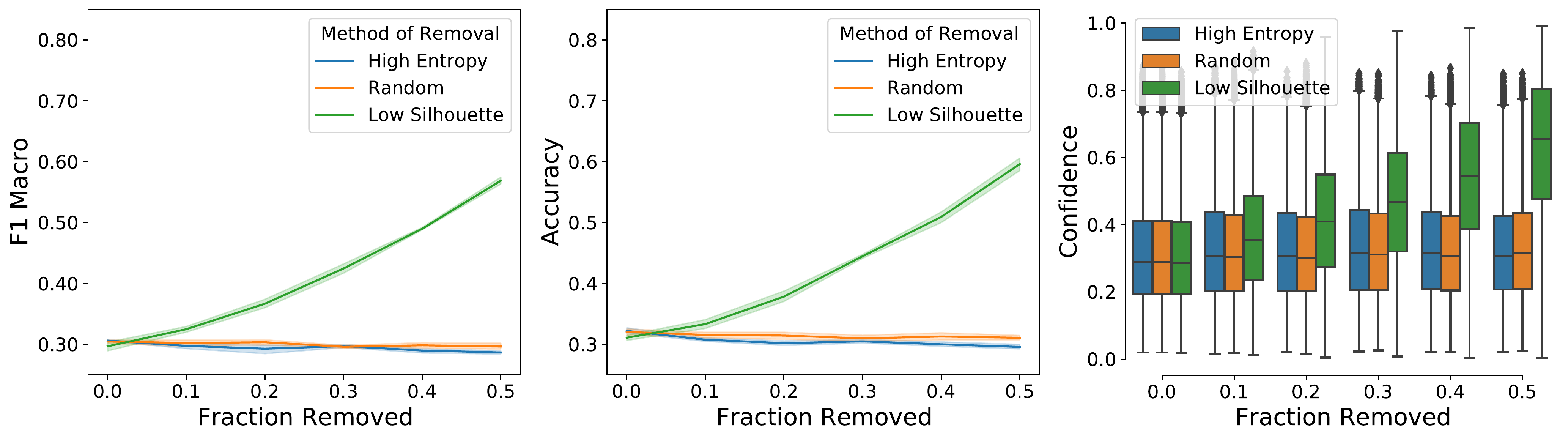}
        \caption{MFNC dataset}
        \label{fig:results_emfd}
    \end{subfigure}
    \begin{subfigure}{\textwidth}
        \centering
        \includegraphics[width=\textwidth]{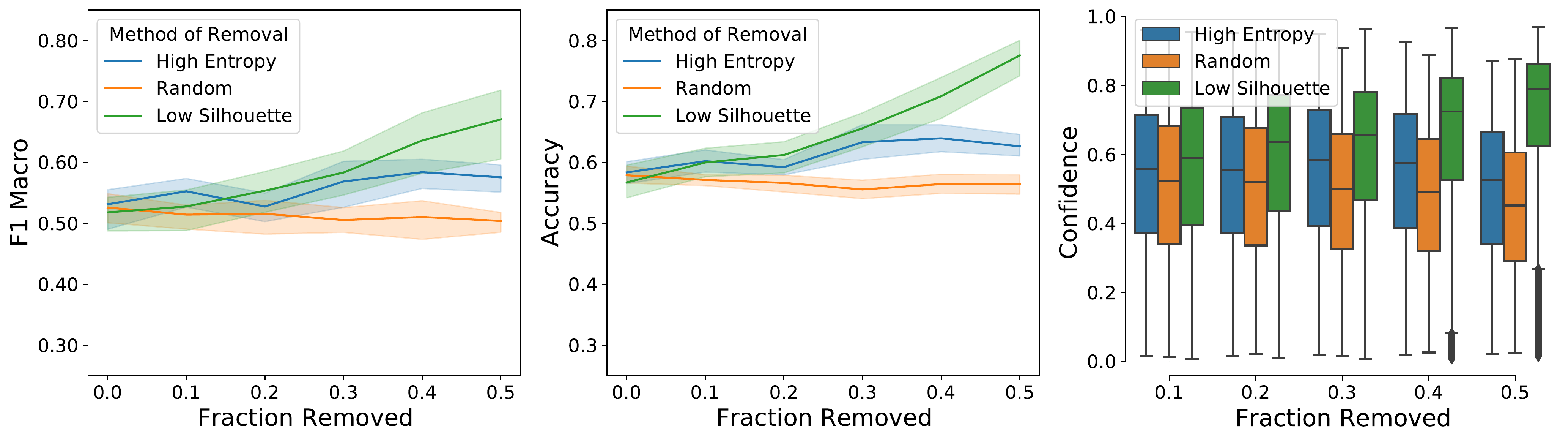}
        \caption{MFRC dataset}
        \label{fig:results_mfrc}
    \end{subfigure}
    \caption{Morality classification after de-noising annotations. Comparison of F1 (left sub-figures), accuracy (middle sub-figures), and distributions of model confidence on instances (right sub-figures) when removing high-entropy instances, low-silhouette judgments, or removing randomly. Removing low-silhouette judgments helps with model's performance on all MFTC, MFNC, and MFRC datasets. However, removing high-entropy instances is only effective on MFTC and doesn't help with learning on MFNC or MFRC due to the high ratio of instances with zero entropy.}
    \label{fig:results}
\end{figure*}

We study the effect of removing noisy annotations on the performance of the trained language model, discarding from each dataset either 1) highest-entropy instances (and all the labels assigned to them), or 2) lowest-silhouette coefficient labels. After filtering the data with a specific ratio, we split the dataset to train and test set. We compare performance to models trained on data from which the same ratio of instances have been removed at random. 
Figure \ref{fig:results} shows performance on the three datasets in terms of the F1 Macro (left subplot), and accuracy (middle) as a function of the fraction of data removed. These measures are aggregated over five runs with different random seeds and calculated on held-out test data. 

Discarding data at random shows a non-increasing trend in the performance of the model. On the other hand, discarding portions of judgments with lowest silhouette coefficients improves the performance of moral classification on all three datasets. These results suggest that our proposed method identifies better-quality ground truth data that helps model performance. 

Although removing highest-entropy instances improves performance of the classifier on MFTC, it does not help improve classification performance on the MFNC dataset and offers only a small improvement on MFRC. This result can be explained by the way each dataset has been collected. In MFTC, each tweet was shown to several annotators and their judgments about all the labels were collected. However, in MFNC, annotators were shown a news document and asked to identify segments of text relevant to a specific moral foundation. For example, annotator $a1$ was asked to highlight any part of document $d$ related to the \textit{care} foundation, but annotator $a2$, $a3$ were asked to highlight portions of $d$ relevant to \textit{authority} and \textit{loyalty}. Their highlighted texts might overlap at sentence $s$, and $s$ at the end will have annotations $<care:1, fairness:0, authority:1, loyalty:1, purity: 0>$. At first glance, it seems there is no agreement on $s$ and its entropy is very high. However, if the annotators were given the opportunity to choose any label, we could have seen more agreement in the distribution of labels. In crowd-sourcing moral annotations, it helps to simplify the task to finding only one moral foundation in a document to reduce confusion \cite{hopp2021extended}. However, this results in a proliferation of instances labeled by only one annotator. The low entropy score will deceptively indicate low disagreement, even though the single annotation could have been made in error.

It is important to understand how the dataset was constructed when choosing between using instance-level entropy and judgment-level silhouette coefficient when preprocessing the data with filtering.

The rightmost panels in Figure~\ref{fig:results} show model confidence as a function of the fraction of data discarded by the entropy or silhouette methods. The figures show that the distribution of the model's confidence on instances shifts toward higher values when we discard labels with lowest silhouette values. As a reminder, higher confidence means that the model is more likely to correctly classify the instances. The same improvement occurs when we discard high-entropy instances in MFTC. Similar to previous results,  removing high-entropy instances in MFNC or MFRC does not improve confidence. Moreover, the distribution of confidence shifts toward lower values or stays the same when we remove instances at random from all datasets. Prior work \cite{swayamdipta-etal-2020-dataset} has shown that if a trained model has low confidence on an instance, it means  that the instance was likely mislabelled. The observed positive shift in the distribution of confidence values suggests that the subset of the data we kept has fewer mislabelled instances.

\section{Conclusions}
In this work we show that auditing annotated data for noise can improve morality classification. We propose two metrics---\textit{entropy} and \textit{silhouette coefficient}---for refining annotated datasets in a pre-processing step, i.e. before training a classifier. The metrics leverage annotator judgments in order to identify instances that are difficult to label or have been mislabeled. We show that removing these instances reduces the noise in the ground truth data, improving classification performance of models trained on the remaining data. 
We also show that refining annotations improves the training dynamics of the model. As a result, the average of confidence increases when the model is trained on the instances that are recognized as less noisy with our proposed metrics. 

We validated our approach on three datasets where multiple annotators were asked to label the moral values expressed in text. Classifying morality is inherently a subjective and difficult task, and the resulting ground truth data from human annotation will naturally contain a large amount of disagreement and noise, which can degrade the performance of single-task models trained on the data. We showed that a classifier trained on refined data, from which the potentially noisy samples have been removed, can learn better models that more accurately recognize new instances of moral foundations. Our approach is not specific to moral annotations and can be applied to other datasets constructed from subjective judgments of multiple annotators.

\section{Limitations and Ethical Considerations}
Other works \cite{denton2021whose, sap2021annotators, waseem-2016-racist, prabhakaran-etal-2021-releasing} have shown annotator demographic features and annotators' life experiences can impact their judgments. 
In this work we focus on single-task classifier models that need an aggregated label (e.g. to majority vote or averaging) for training. However, we encourage future work to design models beyond single-task classifiers in order to overcome the need for aggregating the labels and move to subjective models that can give predictions based on different beliefs, demographics, and backgrounds. 

Also, considering the use-case of a gathered dataset, the removal criteria described in section \ref{sec:method} can be enhanced to make sure we are not removing the samples from specific demographics or groups of people. A possible remedy is to discard weighted portions of data from each demographic group in a way to keep a more balanced subset of data or to prioritize keeping the data gathered by minorities.


\bibliographystyle{IEEEtran}
\bibliography{asonam_bib}
\clearpage

\end{document}